\documentclass[runningheads]{llncs}
\usepackage{graphicx}
\usepackage{multirow}
\usepackage{tikz}
\usepackage{comment}
\usepackage{amsmath,amssymb} % define this before the line numbering.
\usepackage{color,xcolor,colortbl}
\usepackage[pagebackref=false,breaklinks=true,colorlinks,bookmarks=false,citecolor=blue,linkcolor=blue]{hyperref}
\usepackage{caption}
\hypersetup{colorlinks=true,urlcolor=blue}
\usepackage{subfig}
\usepackage{wrapfig}
\usepackage[accsupp]{axessibility}  % Improves PDF readability for those with disabilities.

\usepackage{booktabs}

\usepackage{bm}

\definecolor{lightgray}{gray}{0.95}
\definecolor{color3}{gray}{0.95}
\definecolor{rouse}{rgb}{0.981,0.961,0.941}

\begin{document}
% \linenumbers
\pagestyle{headings}
\mainmatter

\title{Coarse-to-Fine Sparse Transformer for Hyperspectral Image Reconstruction} % Replace with your title
\vspace{-3mm}
%\begin{comment}
%\titlerunning{ECCV-22 submission ID \ECCVSubNumber} 
%\authorrunning{ECCV-22 submission ID \ECCVSubNumber} 
%\author{Anonymous ECCV submission}
%\institute{Paper ID \ECCVSubNumber}
%\end{comment}
%******************

%\begin{comment}
\titlerunning{Coarse-to-Fine Sparse Transformer for Hyperspectral Image Reconstruction}
\author{Yuanhao Cai $^{1,2,*}$, Jing Lin $^{1,2,}$\thanks{Equal Contribution, $\dagger$ Corresponding Author}, Xiaowan Hu $^{1,2}$, Haoqian Wang $^{1,2,\dagger}$, \\ Xin Yuan $^3$, Yulun Zhang $^4$, Radu Timofte $^4$, and Luc Van Gool $^4$ \\
	}
\authorrunning{Yuanhao Cai$^*$ and Jing Lin$^*$ \emph{et al.}}
\institute{$^{1}$ Shenzhen International Graduate School, Tsinghua University, \\ $^2$  Shenzhen Institute of Future Media Technology, \\ $^3$ Westlake University, $^4$ ETH Z\"{u}rich
}
%\end{comment}
%******************
\maketitle

\vspace{-7mm}
\begin{abstract}
Many algorithms have been developed to solve the inverse problem of coded aperture snapshot spectral imaging (CASSI), $i.e.$, recovering the 3D hyperspectral images (HSIs) from a 2D compressive measurement. In recent years, learning-based methods have demonstrated promising performance and dominated the mainstream research direction. However, existing CNN-based methods show limitations in capturing long-range dependencies and non-local self-similarity. Previous Transformer-based methods densely sample tokens, some of which are uninformative, and calculate the multi-head self-attention (MSA) between some tokens that are unrelated in content. This does not fit the spatially sparse nature of HSI signals and limits the model scalability. In this paper, we propose a novel Transformer-based method, coarse-to-fine sparse Transformer (CST), firstly embedding HSI sparsity into deep learning for HSI reconstruction. In particular, CST uses our proposed spectra-aware screening mechanism (SASM) for  $coarse~patch$ $selecting$. Then the selected patches are fed into our customized spectra-aggregation hashing multi-head self-attention (SAH-MSA) for $fine~pixel$ $clustering$ and self-similarity capturing. Comprehensive experiments show that our CST significantly outperforms state-of-the-art methods while requiring cheaper computational costs. \url{https://github.com/caiyuanhao1998/MST}

\vspace{-2mm}
\end{abstract}

\begin{figure*}[t]
	\begin{center}
		\begin{tabular}[t]{c} \hspace{-3mm}
			\includegraphics[width=1\textwidth]{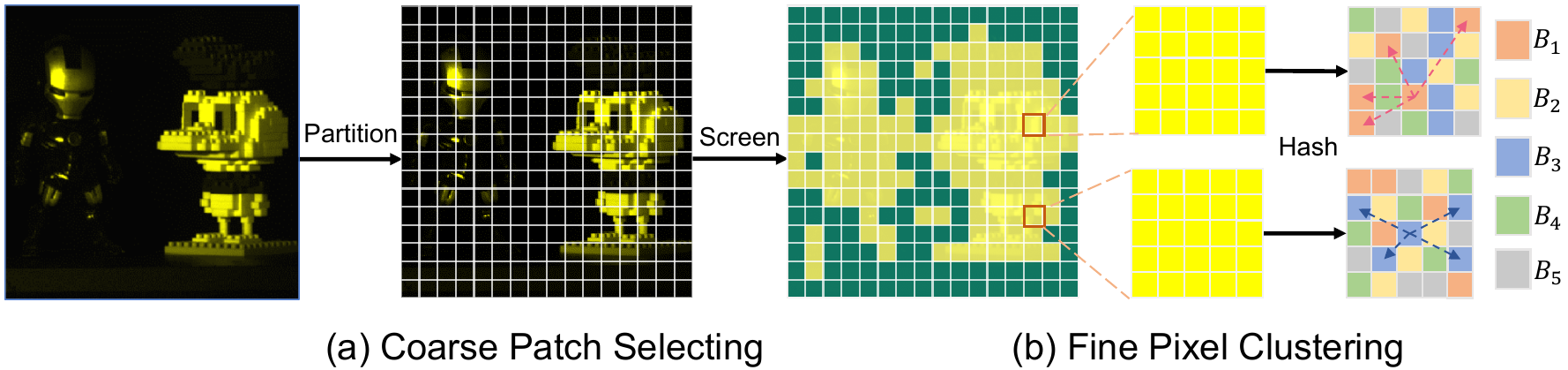}
		\end{tabular}
	\end{center}
	\vspace*{-6mm}
	\caption{\small Diagram of our coarse-to-fine learning scheme. (a) HSIs exhibit high spatial sparsity. The image is firstly partitioned into non-overlapping patches. Then the informative patches (yellow) with HSI signals are screened out and fed into the self-attention mechanism. (b) Tokens closely correlated in content are clustered into the same $bucket$ ($B_1 \sim B_5$). Then the self-attention is calculated inside each $bucket$.}
	\label{fig:teaser}
	\vspace{-5mm}
\end{figure*}

\vspace{-9mm}
\section{Introduction}
\vspace{-3mm}
Hyperspectral images (HSIs), which contain multiple continuous and narrow spectral bands, can provide more detailed information of the captured scene than normal RGB images. Based on the inherently rich and detailed spectral signatures, HSIs have been widely applied to many computer vision tasks and graphical applications, $e.g.,$  image  classification~\cite{fauvel2012advances,maggiori2017recurrent,zhang2015scene}, object tracking~\cite{fu2016exploiting,ot_1,uzkent2016real,uzkent2017aerial}, remote sensing~\cite{borengasser2007hyperspectral,melgani2004classification,rs_4,yuan2017hyperspectral}, medical imaging~\cite{mi_1,mi_2,mi_3}, $etc$. 

To collect HSI cubes, traditional imaging systems scan the scenes with multiple exposures using 1D or 2D sensors. This imaging process is time-consuming and  limited to static objects~\cite{static}. Thus, conventional imaging systems cannot  capture dynamic scenes. 
Recently, researchers have developed several snapshot compressive imaging (SCI) systems to capture HSIs, where the 3D HSI cube is compressed into a single 2D measurement~\cite{cao2011prism,cao2016computational,manakov2013reconfigurable,wagadarikar2008single}. Among these SCI systems, coded aperture snapshot spectral imaging (CASSI) stands out as a promising solution and has become an active research direction~\cite{gehm2007single,gsm,tsa_net,wagadarikar2008single}. CASSI systems modulate HSI signals at different wavelengths by a coded aperture (physical mask) and then vary the
modulation by a disperser, \emph{i.e.}, to shift the modulated images at diﬀerent wavelengths to diﬀerent spatial locations on the detector plane. Subsequently, a reconstruction algorithm is used to restore the 3D HSI cube from the 2D compressive image, which is a core task in CASSI.

To solve this ill-posed inverse problem, traditional  methods~\cite{liu2018rank,wang2015dual,gap_tv} mainly depend on hand-crafted priors and assumptions. The main drawbacks of these model-based methods are that they need to tweak parameters manually, leading to poor generality and slow reconstruction speed. In recent years, deep learning methods have shown the potential to speed up the reconstruction and improve restoration quality for natural images~\cite{pngan,msfn,p3an,poan,rdn,restormer,cycleisp,mirnet,mprnet,rcan,rnan}. Hence, convolution neural networks (CNNs) have been used to learn the underlying mapping function from the 2D compressive measurement to the 3D HSI signal. Nevertheless, these CNN-based methods yield impressive results but show limitations in capturing non-local self-similarity and long-range dependencies. 

In the past few years, the natural language processing (NLP) model  Transformer~\cite{vaswani2017attention} has gained much popularity and achieved great success in computer vision. Transformer provides a powerful model that excels at exploring global inter-dependence between different regions to alleviate the constraints of CNN-based methods. Nonetheless, directly applying vision Transformers to HSI reconstruction encounters two main issues that cannot be ignored. \textbf{Firstly}, the HSI signals exhibit high spatial sparsity as shown in Fig.~\ref{fig:teaser} (a). Some dark regions are almost uninformative. However,  previous local~\cite{liu2021swin} or global~\cite{global_msa} Transformers process all spatial pixel vectors inside non-overlapping windows or global images into tokens without screening and then feed the tokens into the multi-head self-attention (MSA) mechanism. Many regions with limited information are sampled, which dramatically degrades the model efficiency and limits the reconstruction performance. \textbf{Secondly}, previous Transformers linearly project all the tokens into $query$, $key$, and $value$, and then perform matrix multiplication for calculating MSA without clustering. Yet, some of the tokens are not related in content. Attending to all these tokens at once lowers down the cost-effectiveness of model and may easily lead to  over-smooth results~\cite{xiangtl_gald}. \textbf{Besides}, the computational complexity of global Transformer~\cite{global_msa} is quadratic to the spatial dimensions, which is nontrivial and sometimes unaffordable. %The local Transformer reduces the computational cost by constraining the MSA inside position-specific windows but suffers from limited receptve fields. 
MST~\cite{mst} calculates MSA along the spectral dimension, thus circumventing the HSI spatial sparsity. 

Hence, how to combine HSI sparsity with learning-based algorithms still remains under-explored. This work aims to investigate this problem and cope with the  limitations of existing CNN-based and Transformer-based methods.

In this paper, we propose a novel method,  coarse-to-fine sparse Transformer (CST), for HSI reconstruction. Our CST composes two key techniques. \textbf{Firstly}, due to the large variation in HSI informativeness of spatial regions, we propose a spectra-aware screening mechanism (SASM) for \emph{coarse patch selecting}. To be specific, in Fig.~\ref{fig:teaser} (a), our SASM  partitions the image into non-overlapping patches and then detects the patches that are informative of HSI representations. Subsequently, only the detected patches (yellow) are fed into the self-attention mechanisms to decrease the inefficient calculation of uninformative regions (green) and promote the model cost-effectiveness. \textbf{Secondly}, instead of using all projected tokens at once like previous Transformers, we aim to calculate self-attention of tokens that are closely related in content. Toward  this end, we customize spectra-aggregation hashing multi-head self-attention (SAH-MSA) for \emph{fine pixel clustering} as shown in Fig.~\ref{fig:teaser} (b). SAH-MSA learns to cluster tokens into different groups (termed $buckets$ in this paper) by searching similar elements that produce the max inner product. Tokens inside each $bucket$ are considered closely related in content. Then the MSA operation is applied within each $bucket$. \textbf{Finally}, with the proposed techniques, we enable a coarse-to-fine learning scheme that embeds the HSI spatial sparsity into learning-based methods. We establish a series of small-to-large CST families that outperform state-of-the-art (SOTA) methods while requiring much cheaper computational costs.

The main contributions of this work can be summarized as follows:
\begin{itemize}
	%\vspace{-3mm}
	\item We propose a novel Transformer-based method, CST, for HSI reconstruction. To the best of our knowledge, it is the first attempt to embed the HSI spatial sparsity nature into learning-based algorithms for this task. 
	%\vspace{-1.25mm}
	\item We present SASM to locate informative regions with HSI signals.
	%\vspace{-1.25mm}
	\item We customize SAH-MSA to capture interactions of closely related patterns.
	%\vspace{-5.25mm}
	\item Our CST with much lower computational complexity significantly surpasses SOTA algorithms on all scenes in simulation. Moreover, our CST yields more visually pleasant results than existing methods in real HSI restoration. 
\end{itemize}

% \begin{figure*}[t]
% 	\begin{center}
% 		\begin{tabular}[t]{c} \hspace{-2mm}
% 			\includegraphics[width=0.95\textwidth]{figures/teacher_fig.pdf}
% 		\end{tabular}
% 	\end{center}
% 	\vspace*{-6.5mm}
% 	\caption{\small Illustration of the proposed method. Our Spectra-Aware Clustering Sparse Attention (SCSA) is motivated by the HSI characteristics that the representations of HSI is highly sparse. (a) SCAB is composed of  two layer normalization, a LCA, and a Feed-Forward Network (FFN). Only the patches with top-k largest sparsity values will be refined by the LCA. (b) LCA groups the elements into clusters based on the hash codes and computes the self-attention within each cluster, allowing the query to attend to the highly-related keys. The shade of color represents the number of hash code.}
% 	\label{fig:techer_fig}
% 	\vspace{-1mm}
% \end{figure*}

\vspace{-3mm}
\section{Related Work}
\vspace{-2mm}
\subsection{Hyperspectral Image Reconstruction}
\vspace{-1mm}
Conventional HSI reconstruction methods~\cite{twist,gradient,desci,liu2018rank,wang2015dual,gap_tv} rely on hand-crafted image priors. For instance, gradient projection~\cite{gradient} algorithms are exploited to handle the HSI sparseness. In addition, total variation regularizers are employed by GAP-TV~\cite{gap_tv} while the low-rank property and non-local self-similarity are used in DeSCI~\cite{desci}. Nonetheless, these traditional model-based methods suffer from low reconstruction speed and poor generalization ability. Recently, CNNs have been used to solve the inverse problem of spectral SCI. These CNN-based algorithms can be divided into three categories, \emph{i.e.}, end-to-end (E2E) methods, deep unfolding methods, and plug-and-play (PnP) methods. E2E algorithms~\cite{mst,fuying-2021-tpami,hdnet,tsa_net,lambda,e2e_1} apply a deep CNN as a powerful model to learn the E2E mapping function of HSI restoration. Deep unfolding methods~\cite{dauhst,fu2021bidirectional,gsm,admm-net,gapnet,dnu} employ multi-stage CNNs trained to map the measurements into the desired signal. Each stage contains two parts, \emph{i.e.}, linear projection and passing the signal through a CNN functioning as a denoiser. PnP methods~\cite{pnp_3,pnp_1,pnp_2} plug pre-trained CNN denoisers into model-based methods to solve the HSI reconstruction problem. Nonetheless, these CNN-based algorithms show limitations in capturing long-range spatial dependencies and modeling the non-local self-similarity. Besides, the sparsity property of HSI representations is not well addressed, posing a low-efficiency problem  to HSI reconstruction models. 

\vspace{-3.5mm}
\subsection{Vision Transformer}
\vspace{-1.5mm}
Transformer~\cite{vaswani2017attention} is proposed for machine translation in NLP. Recently, it has gained much popularity in computer vision because of its superiority in modeling long-range interactions between spatial regions. Vision Transformer has been widely applied in image classification~\cite{tc_2,tc_1,global_msa,xcit,tc_5,tc_7,tc_3,tc_4,tc_6}, object detection~\cite{DETR,dy_detr,to_3,to_5,to_2,to_4,to_1,de_detr}, semantic segmentation~\cite{ts_5,liu2021swin,ts_4,ts_3,ts_2,ts_6,hromer,ts_1}, human pose estimation~\cite{rsn,th_6,th_4,th_2,th_5,th_7,th_3,th_1}, and so on.  Besides high-level vision, Transformer has also been used in image restoration~\cite{mst,vsrt,ipt,rformer,swinir,fgst,wang2022spectral,uformer}. For example, Cai \emph{et al.}~\cite{mst} propose the first Transformer-based model MST for HSI  reconstruction. MST treats spectral maps as tokens and calculates the self-attention along the spectral dimension. In addition, Wang \emph{et al.}~\cite{uformer} propose a U-shaped Transformer, named UFormer, built by the basic blocks of Swin Transformer~\cite{liu2021swin} for natural image restoration.  However, existing Transformers densely sample tokens, some of which corresponding to the regions with limited information, and calculate MSA between some tokens that are unrelated in content. How to embed HSI spatial sparsity into Transformer to boost the model efficiency still remains under-studied. Our work aims to fill this research gap.

\vspace{-3.5mm}
\section{Mathematical Model of CASSI}
\vspace{-1.5mm}
\begin{wrapfigure}{r}{0.51\textwidth}
	\vspace{-12mm}
	\begin{center}
		\includegraphics[width=0.51\textwidth]{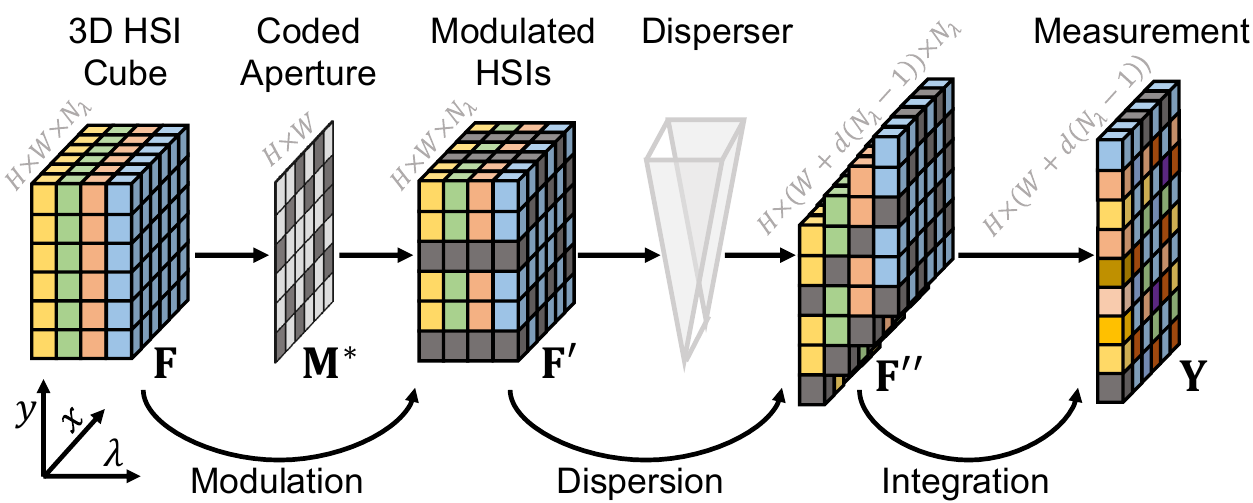}
	\end{center}
	\vspace{-3mm}
	\caption{\small A schematic diagram of CASSI.}
	\vspace{-4mm}
	\label{fig:CASSI}
\end{wrapfigure} 
A schematic diagram of CASSI is shown in Fig.~\ref{fig:CASSI}. We denote the input 3D HSI data cube as $\mathbf{F} \in \mathbb{R}^{H\times W \times N_{\lambda}}$, where $H$, $W$, and $N_\lambda$ refer to the HSI's height, width, and number of wavelengths, respectively. Firstly,  a coded aperture ($i.e.$, a physical mask) $\mathbf{M^*} \in \mathbb{R}^{H\times W}$ is used to modulate $\mathbf{F}$ along the channel dimension:
\vspace{-1.2mm}
\begin{equation}\label{eq:Fprime}
\mathbf{F}' (:,:,n_{\lambda}) = \mathbf{F} (:,:,n_{\lambda}) \odot \mathbf{M^*},
\vspace{-1.2mm}
\end{equation} 
where $\mathbf{F}' \in \mathbb{R}^{H\times W \times N_{\lambda}}$ indicates  the modulated signals, $n_{\lambda} \in [1,\dots, N_{\lambda}]$ indexes the spectral wavelengths, and $\odot$ represents the element-wise product. After undergoing the disperser, $\mathbf{F}'$ becomes tilted and could be treated as  sheared along the $y$-axis. We denote this tilted data cube as $\mathbf{F}'' \in \mathbb{R}^{H\times (W + d(N_{\lambda}-1)) \times N_{\lambda}}$, where $d$ refers to the step of spatial shifting. Suppose $\lambda_c$ is the reference wavelength, which means that $\mathbf{F}'' (:,:,n_{\lambda_c})$ works like an anchor image that is not sheared along the $y$-axis. Then the dispersion can be formulated as
\vspace{-1.5mm}
\begin{equation}\label{eq:Fprime_prime}
\mathbf{F}'' (u,v, n_{\lambda}) = \mathbf{F}'(x, y + d(\lambda_n - \lambda_c), n_{\lambda}),
\vspace{-1.5mm}
\end{equation}
where $(u,v)$ locates the coordinate on the sensoring detector, $\lambda_n$ represents the wavelength of the $n_\lambda$-th channel, and $d(\lambda_n -\lambda_c)$ refers to the spatial shifting offset of the $n_\lambda$-th channel on $\mathbf{F}''$. Eventually, the data cube is compressed into a 2D measurement $\mathbf{Y} \in \mathbb{R}^{H\times (W + d(N_{\lambda}-1))}$ by integrating all the channels as
\vspace{-2.5mm}
\begin{equation}\label{eq:y-discret}
\mathbf{Y} = \sum_{n_{\lambda}=1}^{N_{\lambda}}  \mathbf{F}'' (:,:, n_{\lambda}) +  \mathbf{G}, 
\vspace{-2mm}
\end{equation}
%\vspace{-1.5mm}
where $\mathbf{G}\in \mathbb{R}^{H\times (W + d(N_{\lambda}-1))}$ is the random noise generated during the imaging process. Given the 2D  measurement $\mathbf{Y}$ captured by CASSI, the core task of HSI reconstruction is to restore the 3D HSI data cube $\mathbf{F}$ as mentioned in Eq.~\eqref{eq:Fprime}.

\vspace{-2mm}
\section{Method}
\begin{figure*}[t]
	\begin{center}
		\begin{tabular}[t]{c} \hspace{-2mm}
			\includegraphics[width=1\textwidth]{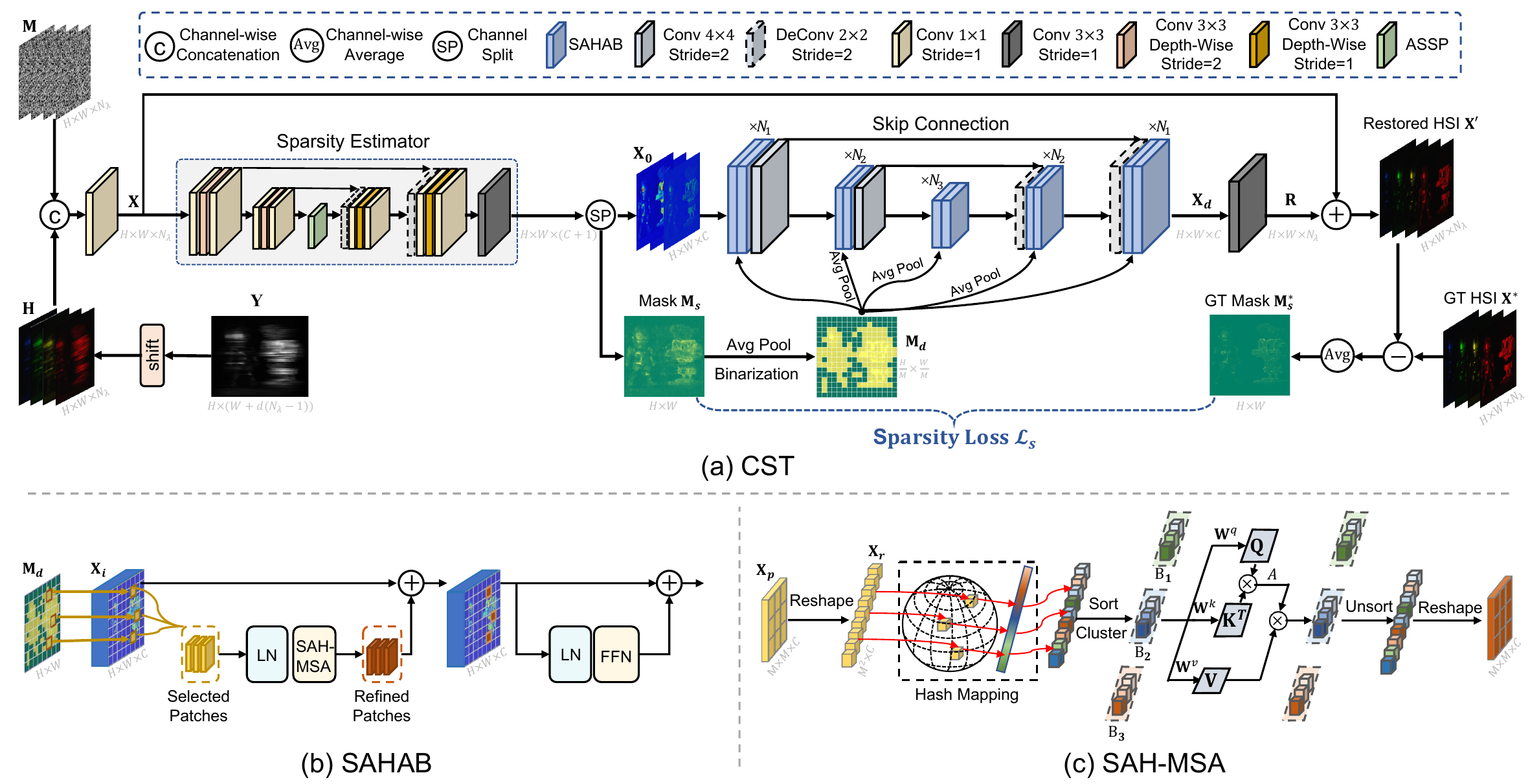}
		\end{tabular}
	\end{center}
	\vspace*{-6.5mm}
	\caption{\small Framework of our method. (a) Spectra-aware screening mechanism (SASM) and the network architecture of CST.  (b) The components of spectra-aware hashing attention block (SAHAB), which is the basic unit of CST. (c) spectra-aggregation hashing multi-head self-attention (SAH-MSA) is the key component embedding in SAHAB.}
	\label{fig:pipeline}
	\vspace{-3.5mm}
\end{figure*}

\vspace{-1mm}
\label{sec:arch}
The overall framework of our coarse-to-fine sparse Transformer (CST) is shown in Fig.~\ref{fig:pipeline}. CST consists of two key components, \emph{i.e.}, spectra-aware screening mechanism (SASM) for \emph{coarse patch selecting} and spectra-aggregation hashing multi-head self-attention (SAH-MSA) for \emph{fine pixel clustering}. Fig.~\ref{fig:pipeline} (a) depicts SASM and the network architecture of CST. Fig.~\ref{fig:pipeline} (b) shows the basic unit of CST, spectra-aware hashing attention block (SAHAB). Fig.~\ref{fig:pipeline} (c) illustrates our SAH-MSA, which is the most important component of SAHAB.

\vspace{-2mm}
\subsection{Network Architecture} 
\vspace{-1mm}
Given a 2D measurement $\mathbf{Y}\in \mathbb{R}^{H\times (W+d(N_\lambda-1))}$, we reverse the dispersion in Eq.~\eqref{eq:Fprime_prime} and shift back $\mathbf{Y}$ to obtain an initialized input signal $\mathbf{H}\in \mathbb{R}^{H\times W \times N_\lambda}$ as
\vspace{-2.2mm}
\begin{equation*}
    \mathbf{H}(x,y,n_\lambda) = \mathbf{Y}(x,y-d(\lambda_n-\lambda_c)).
\label{eq:h_shift_back}
\vspace{-0.3mm}
\end{equation*}
Then $\mathbf{H}$ concatenated with the 3D physical mask $\mathbf{M}\in\mathbb{R}^{H\times W\times N_\lambda}$ (copy the physical mask $\mathbf{M^*}$  $N_\lambda$ times) passes through a $conv$1$\times$1 (convolutional layer with kernel size = 1$\times$1) to generate the initialized feature $\mathbf{X}\in\mathbb{R}^{H\times W\times N_\lambda}$. 
 
\textbf{Firstly}, a sparsity estimator is developed to process $\mathbf{X}$ into a  sparsity mask $\mathbf{M}_s\in \mathbb{R}^{H\times W}$ and shallow feature  $\mathbf{X}_0\in\mathbb{R}^{H\times W\times C}$. The sparsity estimator is detailed in Sec.~\ref{sec:sasm}. \textbf{Secondly}, the shallow feature $\mathbf{X}_0$  passes through a three-stage symmetric encoder-decoder and is embedded into deep feature $\mathbf{X}_d\in\mathbb{R}^{H\times W\times C}$. The $i$-th stage of encoder or decoder contains $N_i$ SAHABs. As shown in Fig.~\ref{fig:pipeline} (b), SAHAB consists of two layer normalization (LN), an SAH-MSA, and a Feed-Forward Network (FFN). For feature downsampling and upsampling, we adopt strided $conv4\times 4$ and $deconv2\times 2$ layers, respectively. To alleviate the information loss caused by the downsample operation, the encoder features are aggregated with the decoder features via the identity connection. \textbf{Finally}, a $conv3\times3$ is applied to $\mathbf{X}_d$ to produce the residual HSIs $\mathbf{R}\in \mathbb{R}^{H\times W\times N_\lambda}$. Then the  reconstructed HSIs $\mathbf{X'}$ can be obtained by the sum of $\mathbf{R}$ and $\mathbf{X}$ , $i.e.$, $\mathbf{X'}=\mathbf{X}+\mathbf{R}$. 

In our implementation, we set the basic channel $C$ = $N_\lambda$ = 28 to store the HSI information and change the combination ($N_1$,$N_2$,$N_3$) in Fig.~\ref{fig:pipeline} (a) to establish our CST families with small, medium, and large model sizes and computational complexities. They are CST-S (1,1,2), CST-M (2,2,2), and CST-L (2,4,6).

\vspace{-3mm}
\subsection{Spectra-Aware Screening Mechanism}
\vspace{-1mm}
\label{sec:sasm}
We observe that the HSI signal exhibits high sparsity in the spatial dimension. However, the original global Transformer~\cite{global_msa} samples all tokens on the feature map while the window-based local Transformer~\cite{liu2021swin} samples all tokens inside every  non-overlapping window. These Transformers sample many uninformative regions to calculate MSA, which degrades the model efficiency. To cope with this problem, we propose SASM for \emph{coarse patch selecting}, \emph{i.e.}, screening out regions with dense HSI information to produce tokens. In this section, we introduce  SASM in three parts, \emph{i.e.}, sparsity estimator, sparsity loss, and patch selection.
\vspace{-2mm}
\subsubsection{Sparsity Estimator.} In this part, we detail the sparsity  estimator mentioned in Sec.~\ref{sec:arch}. As shown in Fig.~\ref{fig:pipeline} (a), the sparsity estimator adopts a U-shaped structure including a two-stage encoder, an ASSP module~\cite{assp}, and a two-stage decoder. Each stage of the encoder consists of two $conv$1$\times$1 and a strided depth-wise $conv$3$\times$3. Each stage of the decoder contains a strided $deconv$2$\times$2, two $conv$1$\times$1, and a depth-wise $conv$3$\times$3. The sparsity estimator takes the initialized feature $\mathbf{X}$ as the input to produce the shallow feature $\mathbf{X}_0$ and the sparsity mask $\mathbf{M}_s$ that localizes and screens out the informative spatial regions with HSI representations. We achieve this by minimizing our proposed sparsity loss. %between the detected sparsity mask $\mathbf{M}_s$ and ground-truth sparsity mask $\mathbf{M}^*_s \in \mathbb{R}^{H\times W}$.
\vspace{-2mm}
\subsubsection{Sparsity Loss.} To supervise $\mathbf{M}_s$, we need a reference that can tell where the spatially sparse HSI information on the HSI is. Since the background is dark and uninformative, the regions with HSI representations are roughly equivalent to the regions that are hard to reconstruct. This statement can be verified by the visual analysis of sparsity mask in Sec.~\ref{sec:exp_ablation}. Therefore, we design our reference signal $\mathbf{M}^*_s \in \mathbb{R}^{H\times W}$ by averaging the differences between the reconstructed HSIs $\mathbf{X}'$ and the ground-truth HSIs $\mathbf{X}^*$ along the spectral dimension to avoid bias as 
\vspace{-1mm}
\begin{equation}
    \mathbf{M}^*_s=\frac{1}{N_\lambda}\sum_{n_\lambda=1}^{N_\lambda}|\mathbf{X'}(:,:,n_\lambda)-\mathbf{X}^*(:,:,n_\lambda)|.
\label{mask_gt}
\vspace{-1mm}
\end{equation}
Subsequently, our sparsity loss $\mathcal{L}_s$ is constructed as the mean squared error between the predicted sparsity mask $\mathbf{M}_s$ and the reference sparsity mask $\mathbf{M}^*_s$ as 
%\vspace{-2mm}
\begin{equation}
    \mathcal{L}_s=||\mathbf{M}_s-\mathbf{M}^*_s||_2.
\label{sparsity_loss}
%\vspace{-1mm}
\end{equation}
By minimizing $\mathcal{L}_s$, the sparsity estimator is encouraged to detect the foreground hard-to-reconstruct regions with HSI representations. In addition, the overall training objective $\mathcal{L}$ is the weighted sum of $\mathcal{L}_s$ and $\mathcal{L}_2$ loss as
%\vspace{-1mm}
\begin{equation}
\mathcal{L}=\mathcal{L}_2 + \lambda \cdot \mathcal{L}_s = ||\mathbf{X}'-\mathbf{X}^*||_2 + \lambda \cdot ||\mathbf{M}_s-\mathbf{M}^*_s||_2, 
\label{overall_loss}
%\vspace{-1mm}
\end{equation}
where $\mathbf{X}^*$ represents the ground-truth HSIs and $\lambda$ refers  to the hyperparameter that controls the importance balance between $\mathcal{L}_2$ and $\mathcal{L}_s$.

\vspace{-2mm}
\subsubsection{Patch Selection.} Our SASM partitions the feature map into non-overlapping patches at the size of $M \times M$. Then the patches with HSI representations are screened out by the predicted sparsity mask $\mathbf{M}_s$ and fed into SAH-MSA   as shown in Fig.~\ref{fig:pipeline} (b). To be specific, $\mathbf{M}_s$ is firstly downsampled by average pooling and then binarized into  $\mathbf{M}_{d}\in\mathbb{R}^{\frac{H}{M}\times \frac{W}{M}}$. We use a hyperparameter, sparsity ratio $\sigma$, to control the  binarization. More specifically, we select the top $k$ patches with the highest values on the downsampled sparsity mask. $k$ is controlled by $\sigma$ that $k = \lfloor (1 - \sigma) \frac{HW}{M^2} \rfloor$. Each pixel on $\mathbf{M}_{d}$ corresponds to an $M\times M$ patch on the feature map and its 0-1 value classifies whether this patch is screened out. Then $\mathbf{M}_{d}$ is applied to the SAH-MSA of each SAHAB. When $\mathbf{M}_{d}$ is used in the $i$-th stage ($i~>$ 1), an average pooling operation is exploited to downsample $\mathbf{M}_{d}$ into $\frac{1}{2^{i-1}}$ size to match the spatial resolution of the feature map of the $i$-th stage. 

\vspace{-3mm}
\subsection{Spectra-Aggregation Hashing Multi-head Self-Attention.} 
\vspace{-1mm}
Previous Transformers calculate MSA between all the sampled tokens, some of which are even unrelated in content. This may lead to inefficient computation that lowers down the model cost-effectiveness and easily hamper  convergence~\cite{de_detr}. The sparse coding methods~\cite{elad2006image,mei2021image,yang2012coupled,yang2010image,zhao2016reducing} assume that image signals can be represented by a sparse linear combination over dictionary signals. Inspired by this, we propose SAH-MSA for $fine~pixel~clustering$.  SAH-MSA enforces a sparsity constraint on the MSA mechanism. In particular, SAH-MSA only calculates self-attention between tokens that are closely correlated in content, which addresses the limitation of previous Transformers.

Our SAH-MSA learns to cluster tokens into different $buckets$ by searching elements that produce the max inner product. As shown in Fig.~\ref{fig:pipeline} (c), We denote a patch feature map as $\mathbf{X}_p\in \mathbb{R}^{M\times M \times C}$ that is screened out by the sparsity mask. We reshape $\mathbf{X}_p$ into $\mathbf{X}_r\in \mathbb{R}^{N \times C}$, where $N = M \times M$ is the number of elements. Subsequently, we use a hash function to aggregate the information in spectral wise and map a $C$-dimensional element (pixel vector) $\bm{x}\in \mathbb{R}^{C}$ into an integer hash code. We formulate this hash mapping $h:\mathbb{R}^C\to \mathbb{Z}$ as 
%\vspace{-1mm}
\begin{equation}
\begin{aligned}
    h(\bm{x}) = \lfloor \frac{\bm{a} \cdot \bm{x}+b}{r} \rfloor,
\end{aligned}
\label{eq:hash}
%\vspace{-1mm}
\end{equation}
where $r \in \mathbb{R}$ is a constant, $\bm{a}\in\mathbb{R}^C$ and $b \in \mathbb{R}$ are random variables satisfying $\bm{a} = (a_1,a_2,...,a_C)$ with $a_i \sim \mathcal{N}(0,1)$ and $b \sim \mathcal{U}(0,r)$ follows a uniform distribution. Then we sort the elements in $\mathbf{X}_r$ according to their hash codes. The $i$-th sorted element is denoted as $\bm{x}_i \in \mathbb{R}^{C}$. Then we split the elements into $buckets$ as
%\vspace{-1mm}
\begin{equation}
\begin{aligned}
    \mathbf{B}_i = \{\bm{x}_j:im+1\leq j \leq (i+1)m\},
\end{aligned}
\label{eq:cluster}
%\vspace{-1mm}
\end{equation}
where $\mathbf{B}_i$ represents the $i$-th $bucket$. Each $bucket$ has $m$ elements. There are $\frac{M\times M}{m}$ $buckets$ in total. 
With our hash clustering scheme, the closely content-correlated tokens are grouped into the same $bucket$. Therefore, the model can reduce the  computational burden between content-unrelated elements by only applying the MSA operation to the tokens within the same $bucket$. More specifically, for a $query$ element $\bm{q}\in \mathbf{B}_i$, our SAH-MSA can be formulated as
%\vspace{-2mm}
\begin{equation}
\begin{aligned}
\text{SAH-MSA}(\boldsymbol{q},\mathbf{B}_i) 
=\sum_{n=1}^{N}\mathbf{W}_n~\text{head}_n(\boldsymbol{q},\mathbf{B}_i),
\end{aligned}
\label{eq:SAH_MSA}
\end{equation}
where $N$ is the number of attention heads. $\mathbf{W}_n\in \mathbb{R}^{C\times d}$ and $\mathbf{W'}_n \in \mathbb{R}^{d\times C}$ are learnable parameters, where $d=\frac{C}{N}$ denotes the dimension of each head. ${A}_{n\bm{q}\bm{k}}$ and $\text{head}_n$ refer to the attention and output of the $n$-th head, formulated as
%\vspace{-2mm}
\begin{equation}
	\mathbf{A}_{n\boldsymbol{q}\boldsymbol{k}} = \underset{\boldsymbol{k}\in\mathbf{B}_i}{\text{softmax}} (\frac{\boldsymbol{q}^T\mathbf{U}_n^T\mathbf{V}_n\boldsymbol{k}}{\sqrt{d}}),~~~~\text{head}_n(\boldsymbol{q},\mathbf{B}_i) = \sum_{\boldsymbol{k}\in\mathbf{B}_i}\mathbf{A}_{n\boldsymbol{q}\boldsymbol{k}}\mathbf{W'}_n\boldsymbol{k}, 
	\label{eq:ScaledDotProductAttn}
%\vspace{-2mm}
\end{equation}
where $\mathbf{U}_n$ and $\mathbf{V}_n \in \mathbb{R}^{d\times C}$ are learnable parameters. With our hashing scheme, the similar elements are at small possibility to fall into different $buckets$. This probability can be further reduced by conducting  multiple rounds of hashing in parallel~\cite{kitaev2020reformer}. $\mathbf{B}_{i}^r$ denotes the $i$-th $bucket$ of the $r$-th round. Then for each head, the multi-round output  is the weighted sum of each single-round output, \emph{i.e.},
%\vspace{-2mm}
\begin{equation}
\begin{aligned}
\text{head}_n(\boldsymbol{q},\mathbf{B}_i) = \sum_{r=1}^{R}w^r_n~\text{head}_n(\boldsymbol{q},\mathbf{B}_i^r), 
\end{aligned}
\label{eq:multi_round}
%\vspace{-2mm}
\end{equation}
where $R$ refers to the round number and $w^r_n$ represents the weight importance of the $r$-th round in the $n$-th head, which scores the similarity between the $query$ element $\bm{q}$ and the elements belonging to $bucket$ $\mathbf{B}_i^r$. $w^r_n$ can be obtained by
\begin{equation}
w^r_n = \frac{\sum_{\bm{k}\in\mathbf{B}_i^r}A_{n\bm{q}\bm{k}}}{\sum_{\hat{r}=1}^R\sum_{\bm{k}\in\mathbf{B}_i^{\hat{r}}}A_{n\bm{q}\bm{k}}}.
\label{eq:weight}
\end{equation}

%\subsubsection{Discussion.} Our spectra-aware hashing sparse Transformer (SHT) is proposed based on the sparsity characteristic of HSI signals and the sparsity are reflected at two aspects: Firstly, instead of performing the self-attention in the whole feature map, we propose a spectra-aware mechanism to sparsely select important patches and only refine these patches with our SAH. Secondly, during performing self-attention in the selected patches, we only attend to the spatially sparse yet highly-related elements based on SAH. With these sparsity constraints, our model are extremely efficient as demonstrated in the next section.

%Our SAH-MSA is an improved version of the original hashing attention~\cite{kitaev2020reformer,mei2021image} and the differences are summarized as follows: Firstly, our hashing algorithm is performed within a local patch scope instead of global scope. This manually introduced constraint can greatly facilitate convergence and lead to a better reconstruction result as shown in our experiment. Secondly, we use the SAH to cluster the input elements before mapping them into query and key embedding. In this way, the computational cost of the hashing function is reduced by half, leading to a more efficient model.

\begin{table*}[t]
	\renewcommand{\arraystretch}{1.0}
	\caption{ Comparisons of Params, FLOPS, PSNR (upper entry in each cell), and SSIM (lower entry in each cell) of different methods on 10 simulation scenes (S1$\sim$S10). Best results are in bold. * denotes setting the sparsity ratio to 0.}
	\vspace{2mm}
	\newcommand{\tabincell}[2]{\begin{tabular}{@{}#1@{}}#2\end{tabular}}
	% \caption{SSIM values by different algorithms on 10 synthetic data.}
	\centering
	\resizebox{\textwidth}{!}
	{
		\centering
		% \begin{tabular}{c|c|c|c|c|c|c|>{\columncolor{lightgray}}c}
		\begin{tabular}{cccccccccccccc}
			\toprule[0.2em]
			\rowcolor{lightgray}
			Algorithms
			&~~Params~~
			&~GFLOPS~
			& ~~~~S1~~~~
			& ~~~~S2~~~~
			& ~~~~S3~~~~
			& ~~~~S4~~~~
			& ~~~~S5~~~~
			& ~~~~S6~~~~
			& ~~~~S7~~~~
			& ~~~~S8~~~~
			& ~~~~S9~~~~
			& ~~~~S10~~~~
			& ~~~~Avg~~~~
			\\
			\midrule[0.1em]
			TwIST \cite{twist}
			& - 
			& -
			&\tabincell{c}{25.16\\0.700}
			&\tabincell{c}{23.02\\0.604}
			&\tabincell{c}{21.40\\0.711}
			&\tabincell{c}{30.19\\0.851}
			&\tabincell{c}{21.41\\0.635}
			&\tabincell{c}{20.95\\0.644}
			&\tabincell{c}{22.20\\0.643}
			&\tabincell{c}{21.82\\0.650}
			&\tabincell{c}{22.42\\0.690}
			&\tabincell{c}{22.67\\0.569}
			&\tabincell{c}{23.12\\0.669}
			\\
			\midrule[0.1em]
			GAP-TV \cite{gap_tv}
			& - 
			& -
			&\tabincell{c}{26.82\\0.754}
			&\tabincell{c}{22.89\\0.610}
			&\tabincell{c}{26.31\\0.802}
			&\tabincell{c}{30.65\\0.852}
			&\tabincell{c}{23.64\\0.703}
			&\tabincell{c}{21.85\\0.663}
			&\tabincell{c}{23.76\\0.688}
			&\tabincell{c}{21.98\\0.655}
			&\tabincell{c}{22.63\\0.682}
			&\tabincell{c}{23.10\\0.584}
			&\tabincell{c}{24.36\\0.669}
			\\
			\midrule[0.1em]
			DeSCI \cite{desci}
			& - 
			& -
			&\tabincell{c}{27.13\\0.748}
			&\tabincell{c}{23.04\\0.620}
			&\tabincell{c}{26.62\\0.818}
			&\tabincell{c}{34.96\\0.897}
			&\tabincell{c}{23.94\\0.706}
			&\tabincell{c}{22.38\\0.683}
			&\tabincell{c}{24.45\\0.743}
			&\tabincell{c}{22.03\\0.673}
			&\tabincell{c}{24.56\\0.732}
			&\tabincell{c}{23.59\\0.587}
			&\tabincell{c}{25.27\\0.721}
			\\
			\midrule[0.1em]
			$\lambda$-net \cite{lambda}
			& 62.64M
			& 117.98
			&\tabincell{c}{30.10\\0.849}
			&\tabincell{c}{28.49\\0.805}
			&\tabincell{c}{27.73\\0.870}
			&\tabincell{c}{37.01\\0.934}
			&\tabincell{c}{26.19\\0.817}
			&\tabincell{c}{28.64\\0.853}
			&\tabincell{c}{26.47\\0.806}
			&\tabincell{c}{26.09\\0.831}
			&\tabincell{c}{27.50\\0.826}
			&\tabincell{c}{27.13\\0.816}
			&\tabincell{c}{28.53\\0.841}
			\\
			\midrule[0.1em]
			HSSP \cite{hssp}
			& - 
			& -
			&\tabincell{c}{31.48\\0.858}
			&\tabincell{c}{31.09\\0.842}
			&\tabincell{c}{28.96\\0.823}
			&\tabincell{c}{34.56\\0.902}
			&\tabincell{c}{28.53\\0.808}
			&\tabincell{c}{30.83\\0.877}
			&\tabincell{c}{28.71\\0.824}
			&\tabincell{c}{30.09\\0.881}
			&\tabincell{c}{30.43\\0.868}
			&\tabincell{c}{28.78\\0.842}
			&\tabincell{c}{30.35\\0.852}
			\\
			\midrule[0.1em]
			DNU \cite{dnu}
			& 1.19M
			& 163.48
			&\tabincell{c}{31.72\\0.863}
			&\tabincell{c}{31.13\\0.846}
			&\tabincell{c}{29.99\\0.845}
			&\tabincell{c}{35.34\\0.908}
			&\tabincell{c}{29.03\\0.833}
			&\tabincell{c}{30.87\\0.887}
			&\tabincell{c}{28.99\\0.839}
			&\tabincell{c}{30.13\\0.885}
			&\tabincell{c}{31.03\\0.876}
			&\tabincell{c}{29.14\\0.849}
			&\tabincell{c}{30.74\\0.863}
			\\
			\midrule[0.1em]
			DIP-HSI \cite{self}
			& 33.85M
			& 64.42
			&\tabincell{c}{32.68\\0.890}
			&\tabincell{c}{27.26\\0.833}
			&\tabincell{c}{31.30\\0.914}
			&\tabincell{c}{40.54\\0.962}
			&\tabincell{c}{29.79\\0.900}
			&\tabincell{c}{30.39\\0.877}
			&\tabincell{c}{28.18\\0.913}
			&\tabincell{c}{29.44\\0.874}
			&\tabincell{c}{34.51\\0.927}
			&\tabincell{c}{28.51\\0.851}
			&\tabincell{c}{31.26\\0.894}
			\\
			\midrule[0.1em]
			TSA-Net \cite{tsa_net}
			& 44.25M
			& 110.06
			&\tabincell{c}{32.03\\0.892}
			&\tabincell{c}{31.00\\0.858}
			&\tabincell{c}{32.25\\0.915}
			&\tabincell{c}{39.19\\0.953}
			&\tabincell{c}{29.39\\0.884}
			&\tabincell{c}{31.44\\0.908}
			&\tabincell{c}{30.32\\0.878}
			&\tabincell{c}{29.35\\0.888}
			&\tabincell{c}{30.01\\0.890}
			&\tabincell{c}{29.59\\0.874}
			&\tabincell{c}{31.46\\0.894}
			\\
			\midrule[0.1em]
			DGSMP \cite{gsm}
			& 3.76M
			& 646.65
			&\tabincell{c}{33.26\\0.915}
			&\tabincell{c}{32.09\\0.898}
			&\tabincell{c}{33.06\\0.925}
			&\tabincell{c}{40.54\\0.964}
			&\tabincell{c}{28.86\\0.882}
			&\tabincell{c}{33.08\\0.937}
			&\tabincell{c}{30.74\\0.886}
			&\tabincell{c}{31.55\\0.923}
			&\tabincell{c}{31.66\\0.911}
			&\tabincell{c}{31.44\\0.925}
			&\tabincell{c}{32.63\\0.917}
			\\
			\midrule[0.1em]
			HDNet \cite{hdnet}
			& 2.37M
			& 154.76
			&\tabincell{c}{35.14\\0.935}
			&\tabincell{c}{35.67\\0.940}
			&\tabincell{c}{36.03\\0.943}
			&\tabincell{c}{42.30\\0.969}
			&\tabincell{c}{32.69\\0.946}
			&\tabincell{c}{34.46\\0.952}
			&\tabincell{c}{33.67\\0.926}
			&\tabincell{c}{32.48\\0.941}
			&\tabincell{c}{34.89\\0.942}
			&\tabincell{c}{32.38\\0.937}
			&\tabincell{c}{34.97\\0.943}
			\\
			\midrule[0.1em]
			MST-S \cite{mst}
			& \bf 0.93M
			& 12.96
			&\tabincell{c}{34.71\\0.930}
			&\tabincell{c}{34.45\\0.925}
			&\tabincell{c}{35.32\\0.943}
			&\tabincell{c}{41.50\\0.967}
			&\tabincell{c}{31.90\\0.933}
			&\tabincell{c}{33.85\\0.943}
			&\tabincell{c}{32.69\\0.911}
			&\tabincell{c}{31.69\\0.933}
			&\tabincell{c}{34.67\\0.939}
			&\tabincell{c}{31.82\\0.926}
			&\tabincell{c}{34.26\\0.935}
			\\
			\midrule[0.1em]
			MST-M \cite{mst}
			& 1.50M
			& 18.07
			&\tabincell{c}{35.15\\0.937}
			&\tabincell{c}{35.19\\0.935}
			&\tabincell{c}{36.26\\0.950}
			&\tabincell{c}{\bf{42.48}\\0.973}
			&\tabincell{c}{32.49\\0.943}
			&\tabincell{c}{34.28\\0.948}
			&\tabincell{c}{33.29\\0.921}
			&\tabincell{c}{32.40\\0.943}
			&\tabincell{c}{35.35\\0.942}
			&\tabincell{c}{32.53\\0.935}
			&\tabincell{c}{34.94\\0.943}
			\\
			\midrule[0.1em]
			MST-L \cite{mst}
			& 2.03M
			& 28.15
			&\tabincell{c}{35.40\\0.941}
			&\tabincell{c}{35.87\\0.944}
			&\tabincell{c}{36.51\\0.953}
			&\tabincell{c}{42.27\\0.973}
			&\tabincell{c}{32.77\\0.947}
			&\tabincell{c}{34.80\\0.955}
			&\tabincell{c}{33.66\\0.925}
			&\tabincell{c}{32.67\\0.948}
			&\tabincell{c}{35.39\\0.949}
			&\tabincell{c}{32.50\\0.941}
			&\tabincell{c}{35.18\\0.948}
			\\
			\midrule[0.1em]
			%\rowcolor{lightgray}
			\rowcolor{rouse}
			\bf CST-S 
			& 1.20M
			& \bf 11.67
			&\tabincell{c}{34.78\\0.930}
			&\tabincell{c}{34.81\\0.931}
			&\tabincell{c}{35.42\\0.944}
			&\tabincell{c}{41.84\\0.967}
			&\tabincell{c}{32.29\\0.939}
			&\tabincell{c}{34.49\\0.949}
			&\tabincell{c}{33.47\\0.922}
			&\tabincell{c}{32.89\\0.945}
			&\tabincell{c}{34.96\\0.944}
			&\tabincell{c}{32.14\\0.932}
			&\tabincell{c}{34.71\\0.940}
			\\
			\midrule[0.1em]
			\rowcolor{rouse}
			\bf CST-M 
			& 1.36M
			& 16.91
			&\tabincell{c}{35.16\\0.938}
			&\tabincell{c}{35.60\\0.942}
			&\tabincell{c}{36.57\\0.953}
			&\tabincell{c}{42.29\\0.972}
			&\tabincell{c}{32.82\\0.948}
			&\tabincell{c}{35.15\\0.956}
			&\tabincell{c}{33.85\\0.927}
			&\tabincell{c}{33.52\\0.952}
			&\tabincell{c}{35.28\\0.946}
			&\tabincell{c}{32.84\\0.940}
			&\tabincell{c}{35.31\\0.947}
			\\
			\midrule[0.1em]
			\rowcolor{rouse}
			\bf CST-L 
			& 3.00M
			& 27.81
			&\tabincell{c}{35.82\\0.947}
			&\tabincell{c}{36.54\\0.952}
			&\tabincell{c}{37.39\\0.959}
			&\tabincell{c}{42.28\\0.972}
			&\tabincell{c}{33.40\\0.953}
			&\tabincell{c}{35.52\\0.962}
			&\tabincell{c}{34.44\\0.937}
			&\tabincell{c}{33.83\\0.959}
			&\tabincell{c}{35.92\\0.951}
			&\tabincell{c}{\bf{33.36}\\\bf{0.948}}
			&\tabincell{c}{35.85\\0.954}
			\\
			\midrule[0.1em]
			\rowcolor{rouse}
			\bf CST-L$^*$
			& 3.00M
			& 40.10
			&\tabincell{c}{\bf{35.96}\\\bf{0.949}}
			&\tabincell{c}{\bf{36.84}\\\bf{0.955}}
			&\tabincell{c}{\bf{38.16}\\\bf{0.962}}
			&\tabincell{c}{42.44\\\bf{0.975}}
			&\tabincell{c}{\bf{33.25}\\\bf{0.955}}
			&\tabincell{c}{\bf{35.72}\\\bf{0.963}}
			&\tabincell{c}{\bf{34.86}\\\bf{0.944}}
			&\tabincell{c}{\bf{34.34}\\\bf{0.961}}
			&\tabincell{c}{\bf{36.51}\\\bf{0.957}}
			&\tabincell{c}{33.09\\0.945}
			&\tabincell{c}{\bf{36.12}\\\bf{0.957}}
			\\
			\bottomrule[0.2em]
		\end{tabular}
	}
	\vspace{-4mm}
	\label{Tab:simu}
\end{table*}

\vspace{-4mm}
\section{Experiment}
\vspace{-2mm}
\label{sec:exp}
\subsection{Experiment Settings}
\vspace{-1mm}
The same with TSA-Net~\cite{tsa_net}, 28 wavelengths from 450 nm to 650 nm are derived by spectral interpolation manipulation for simulation and real experiments.

\noindent \textbf{Synthetic Data.} Two HSI datasets, CAVE~\cite{cave} and KAIST~\cite{kaist}, are adopted for simulation experiments. CAVE contains 32 HSIs with spatial size 512$\times$512. KAIST is composed of 30 HSIs with spatial size 2704$\times$3376. Similar to \cite{mst,gsm,tsa_net}, CAVE is used for training and 10 scenes from KAIST are  selected for testing. 

\noindent \textbf{Real Data.} We adopt the real HSI dataset collected by TSA-Net~\cite{tsa_net}.

\noindent \textbf{Evaluation Metrics.} We use peak signal-to-noise ratio (PSNR) and structural similarity (SSIM)~\cite{ssim} as metrics to evaluate HSI reconstruction methods.

\noindent \textbf{Implementation Details.} Our CST models are implemented by Pytorch. They are trained with Adam~\cite{adam} optimizer ($\beta_1$ = 0.9 and $\beta_2$ = 0.999) using Cosine Annealing scheme~\cite{cosine} for 500 epochs. The learning rate is initially set to 4$\times$10$^{-4}$. In simulation experiments, patches at the spatial size of 256$\times$256 are randomly cropped from the 3D HSI cubes with 28 channels as training samples. For real HSI  reconstruction, we set the spatial size of patches to 660$\times$660  with the same size of the real physical mask. We set the shifting step $d$ in the dispersion to 2. After the mask modulation, the image cube is sheared with an accumulative two-pixel step. Hence, the spatial sizes of measurements are 256$\times$310 and 660$\times$714 in simulation and real experiments. The batch size is set to 5. $r$ and $m$ in Eq.~\eqref{eq:hash} and \eqref{eq:cluster} are set to 1 and 64.  The training data is augmented with random  rotation and flipping. All CST models are trained and tested on a single RTX 3090 GPU. 

\begin{figure*}[t]
	\begin{center}
		\begin{tabular}[t]{c} \hspace{-2mm}
			\includegraphics[width=1.0\textwidth]{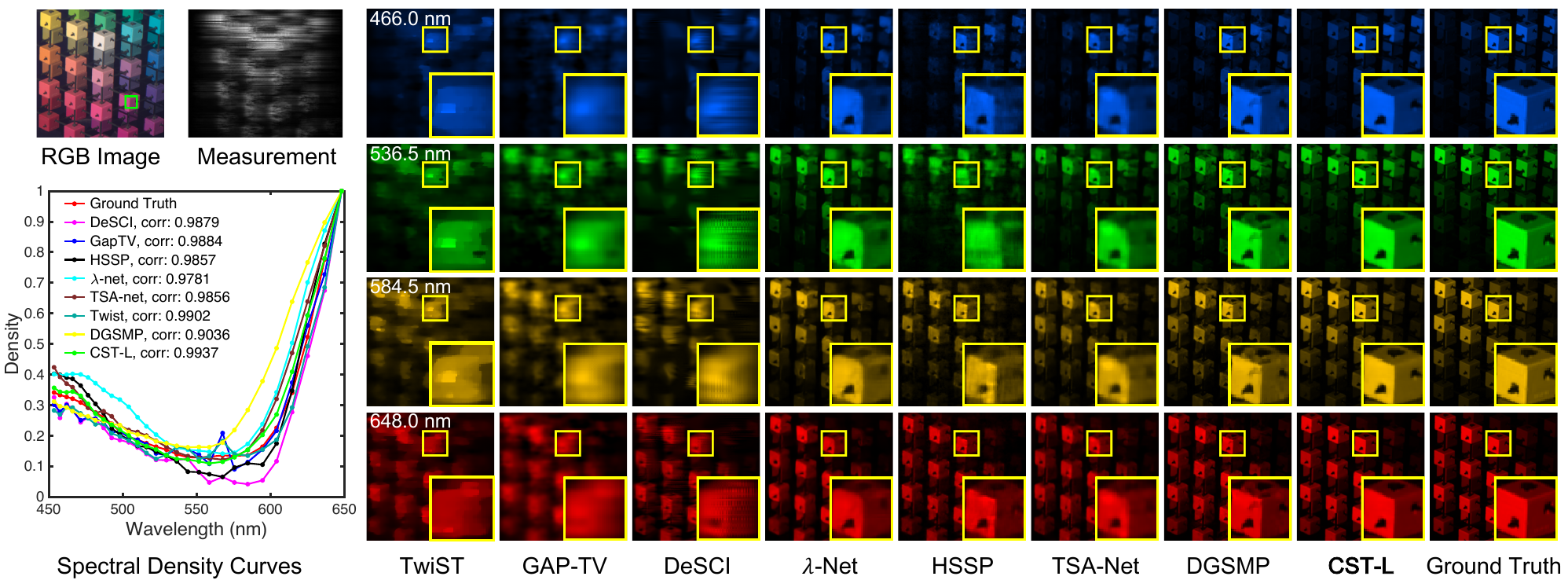}
		\end{tabular}
	\end{center}
	\vspace*{-6mm}
	\caption{\small Reconstructed simulation HSI comparisons of  \emph{Scene} 2 with 4 out of 28 spectral channels. 7 SOTA methods and our CST-L are included. The spectral curves (bottom-left) are corresponding to the selected green box of the RGB image. Please zoom in. }
	\label{fig:simulation}
	\vspace{-4mm}
\end{figure*}

\vspace{-3mm}
\subsection{Quantitative Results}
\vspace{-1mm}
\begin{wrapfigure}{r}{0.41\textwidth}
	\vspace{-12mm}
	\begin{center}
		\includegraphics[width=0.41\textwidth]{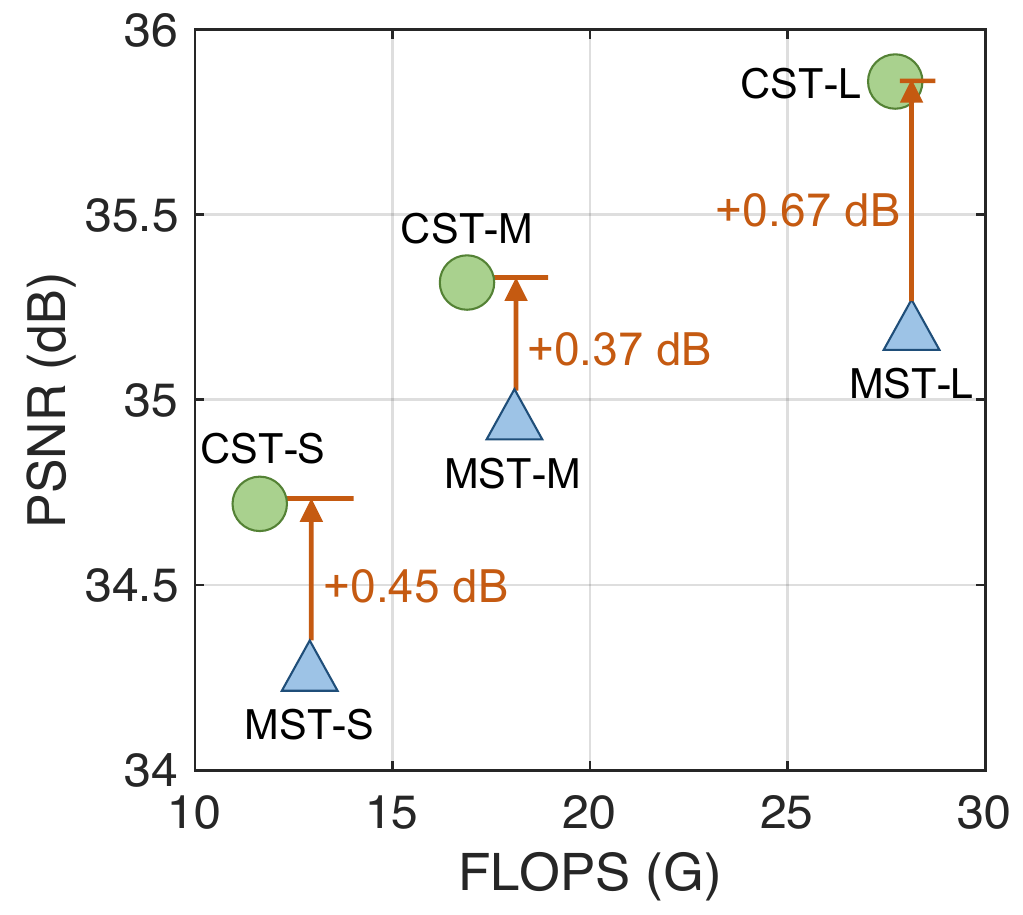}
	\end{center}
	\vspace{-6mm}
	\caption{\small CST \emph{vs}. MST.}
	\vspace{-7mm}
	\label{fig:cst_mst}
\end{wrapfigure} 
We compare the Params, FLOPS, PSNR, and SSIM of our CST and other SOTA methods, including three model-based methods (TwIST~\cite{twist}, GAP-TV~\cite{gap_tv}, and DeSCI~\cite{desci}), six CNN-based methods ($\lambda$-net~\cite{lambda}, HSSP~\cite{hssp}, DNU~\cite{dnu}, PnP-DIP-HSI~\cite{self}, TSA-Net~\cite{tsa_net}, and DGSMP~\cite{gsm}), and a recent Transformer-based method (MST~\cite{mst}). For fairness, we test all these algorithms with the same settings as \cite{mst,gsm}.  The results on 10 simulation scenes are reported in Tab.~\ref{Tab:simu}. 
As can be seen: \textbf{(i)} When we set the sparsity ratio to 0, our best model CST-L$^{*}$ achieves very impressive results, \emph{i.e.}, 36.12 dB in PSNR and 0.957 in SSIM, showing the effectiveness of our method. \textbf{(ii)} Our CST families significantly outperform other SOTA algorithms while requiring cheaper computational costs. Particularly, when compared to the recent best  Transformer-based method MST, our CST-S, CST-M, and CST-L achieve 0.45, 0.37, and 0.67 dB improvements while costing 1.29G, 1.16G, and 0.34G less FLOPS than MST-S, MST-M, and MST-L as shown in Fig.~\ref{fig:cst_mst}. When compared to CNN-based methods, our CST exhibits extreme efficiency advantages. For instance, CST-L outperforms DGSMP, TSA-Net, and $\lambda$-Net by 3.22, 4.39, and 7.32 dB while costing 79.8\% (3.00 / 3.76), 6.8\%, 4.8\% Params and 4.3\% (27.81 / 646.65), 25.3\%, 23.6\% FLOPS. Surprisingly, even our smallest model CST-S surpasses DGSMP, TSA-Net, and $\lambda$-Net by 2.08, 3.25, and 6.18 dB while requiring 31.9\%, 2.7\%, 1.9\% Params and 1.8\%, 10.6\%, 9.9\% FLOPS. These results demonstrate the cost-effectiveness superiority of our  CST. This is mainly because CST embeds the HSI sparsity into the learning-based model, which reduces  the  inefficient computation of less informative dark regions and self-attention between content-unrelated tokens. 

\begin{figure*}[t]
	\begin{center}
		\begin{tabular}[t]{c} \hspace{-2.7mm} %\vspace{-6mm}
			\includegraphics[width=\textwidth]{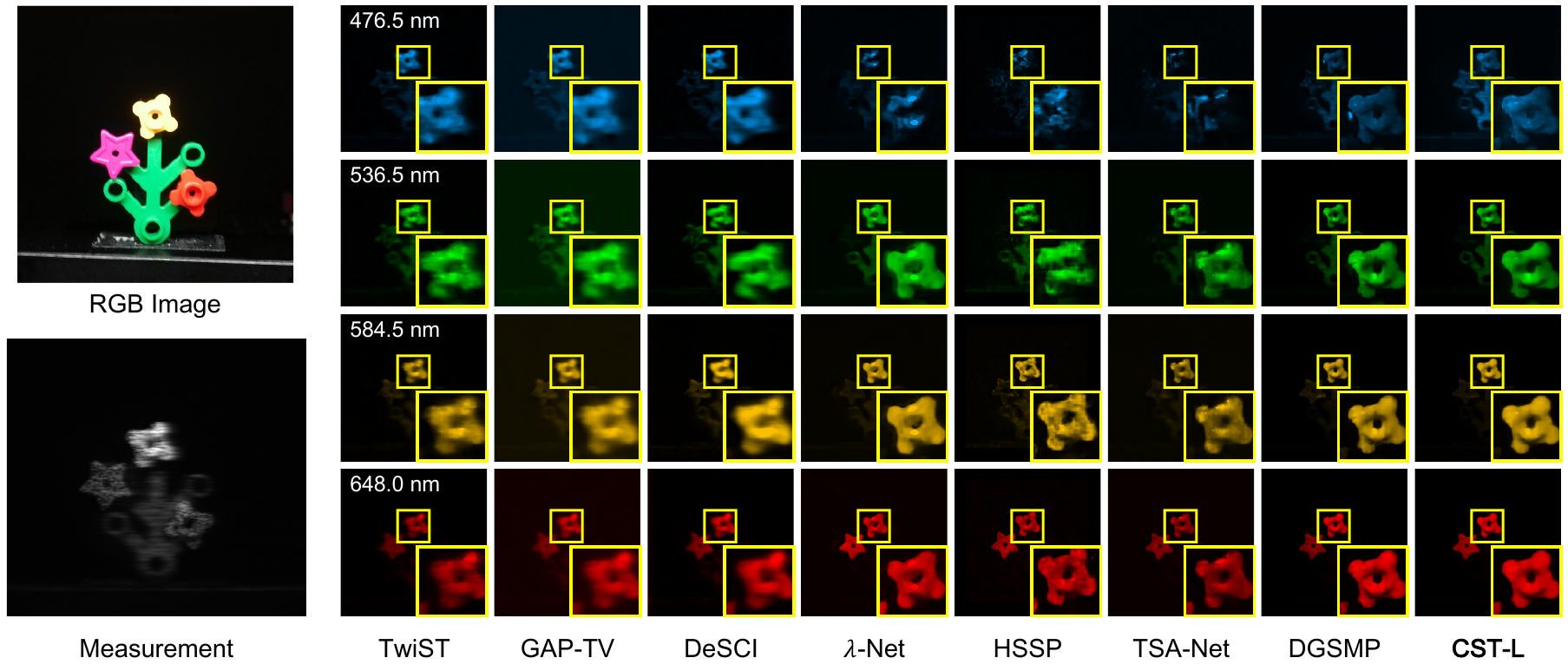}
		\end{tabular}
	\end{center}
	\vspace*{-7mm}
	\caption{\small Reconstructed real HSI comparisons of  \emph{Scene} 1 with 4 out of 28 spectral channels. Seven SOTA algorithms and our CST-L are included. Our CST-L are more favorable to restore detailed contents and remove noise. Zoom in for a better view. }
	\label{fig:real}
	\vspace{-5mm}
\end{figure*}

\vspace{-2mm}
\subsection{Qualitative Results}
\vspace{-1mm}
\subsubsection{Simulation HSI Restoration.} Fig.~\ref{fig:simulation} compares the restored simulation HSIs of our CST-L and seven SOTA algorithms on $Scene$ 2 with 4 out of 28 spectral channels. Please zoom in for better visualization.  It can be observed from the reconstructed HSIs (right) and the zoomed-in patches in the yellow boxes that our CST is effective in producing perceptually pleasant images with more vivid sharp edge details while maintaining the spatial smoothness of the homogeneous regions without introducing artifacts. In contrast, other methods fail to restore fine-grained details. They either achieve over-smooth results sacrificing structural contents and high-frequency details, or generate blotchy textures and chromatic artifacts. Besides, Fig.~\ref{fig:simulation} depicts the spectral density curves (bottom-left) corresponding to the selected region of the green box in the RGB image (top-left). Our curve achieves the highest correlation coefficient with the ground-truth curve. This evidence clearly demonstrates the spectral-dimension consistency reconstruction effectiveness of our proposed CST.

\vspace{-2mm}
\subsubsection{Real HSI Restoration.} We also evaluate our CST in real HSI reconstruction. Following the setting of \cite{mst,gsm,tsa_net},  we re-train our CST-L with all samples of the KAIST and CAVE datasets. To simulate real CASSI, 11-bit shot noise is injected into the measurement during the training procedure. The reconstructed HSI comparisons are depicted in Fig.~\ref{fig:real}. Our CST-L shows significant advantages in fine-grained content restoration and real noise removal.  These results verify the robustness, reliability, and generalization ability of our method.

\begin{table*}[t]%\vspace{-3mm}
	\caption{\small Ablations. Models are trained on CAVE and test on KAIST in simulation. }
	\vspace{1mm}
	% subfloat a - RoIAlign (ResNet-50-C5)
	\subfloat[\small Break-down ablation study. \label{tab:breakdown}]{ 
		\scalebox{0.65}{
			\begin{tabular}{l  c c c}
				%\small
				\toprule
				\rowcolor{color3} Method &~Baseline~ &~+ SAH-MSA~ &~+ SASM~  \\
				\midrule
				PSNR &32.57 &\bf 35.53 &35.31 (\textcolor{blue}{$\downarrow$ 0.60 \%}) \\
				SSIM  &0.906 &\bf 0.948 &0.947 (\textcolor{blue}{$\downarrow$ 0.10 \%})\\
				%\checkmark & &\checkmark &34.02 &0.930 &0.76 &10.02 \\
				Params (M)  &0.51 &1.36 &1.36 ~(\textcolor{blue}{$\downarrow$ 0.00 \%}) \\
				FLOPS (G) &6.40 &24.60 &16.91 (\textcolor{blue}{$\downarrow$ 31.3 \%}) \\
				\bottomrule
	\end{tabular}}}\hspace{2mm}\vspace{0mm}
	% subfloat b - RoIAlign (ResNet-50-C5)
	\subfloat[\small Ablation study of sparse mechanisms. \label{tab:sparse}]{ 
		\scalebox{0.65}{
			\begin{tabular}{l  c c c c}
				%\small
				\toprule
				\rowcolor{color3} Method &~Baseline~ &~Random Sparsity~ &~Uniform Sparsity~ &\bf SASM  \\
				\midrule
				PSNR &32.57 &34.37 &34.33 &\bf 35.31\\
				SSIM  &0.906 &0.937 &0.936 &\bf 0.947\\
				%\checkmark & &\checkmark &34.02 &0.930 &0.76 &10.02 \\
				Params (M)  &0.51 &1.36 &1.36 &1.36\\
				FLOPS (G) &6.40 &16.89 &16.89 &16.91\\
				\bottomrule
	\end{tabular}}}\hspace{4mm}\vspace{0mm}
	% subfloat c - mask representation
	\subfloat[\small Ablation study of self-attention mechanisms.\label{tab:attention}]{
		\scalebox{0.67}{
			\begin{tabular}{l c c c c c c}
				\toprule
				\rowcolor{color3} Method &~Baseline~ &~G-MSA~ &~W-MSA~ &~Swin-MSA~ &~S-MSA~ &\bf SAH-MSA\\
				\midrule
				PSNR &32.57 &35.04 &35.02 &35.12 &35.21 &\bf 35.53 \\
				SSIM &0.906 &0.944 &0.943 &0.945 &0.946 &\bf 0.948 \\
				Params (M)  &0.51 &1.85 &1.85 &1.85 &1.66 &\bf 1.36 \\
				FLOPS (G)  &6.40 &35.58 &24.98 &24.98 &24.74 &\bf 24.60 \\
				\bottomrule
	\end{tabular}}}%\vspace{2mm}
	% subfloat d - RoIAlign (ResNet-50-C5)
	\subfloat[\small Study of clustering scope. \label{tab:cluster}]{ 
		\scalebox{0.67}{
			\begin{tabular}{l  c c c}
				%\small
				\toprule
				\rowcolor{color3} Method &~~Baseline~~ &~~Global~~ &\bf ~Local~  \\
				\midrule
				PSNR &32.57 &35.33 &\bf 35.53 \\
				SSIM  &0.906 &0.946 &\bf 0.948 \\
				%\checkmark & &\checkmark &34.02 &0.930 &0.76 &10.02 \\
				Params (M)  &0.51 &1.36 &1.36 \\
				FLOPS (G) &6.40 &24.60 &24.60 \\
				\bottomrule
	\end{tabular}}}%\hspace{2mm}\vspace{0mm}
	\label{tab:ablations}\vspace{-6mm}
\end{table*}

\subsection{Ablation Study}
\label{sec:exp_ablation} 
We adopt the simulation HSI datasets~\cite{kaist,cave} to conduct ablation studies. The baseline model is derived by removing our SAH-MSA and SASM from CST-M. 
\vspace{-2mm}
\subsubsection{Break-down Ablation.} We firstly perform a break-down ablation to investigate the effect of each component and their interactions. The results are listed in Tab.~\ref{tab:breakdown}. The baseline model yields 32.57 dB in PSNR and 0.906 in SSIM. When SAH-MSA is applied, the  performance gains by 2.96 dB in PSNR and 0.042 in SSIM, showing its significant contribution. When we continue to exploit SASM, the computational cost dramatically declines by 31.3\% (7.69 / 24.60) while the performance only degrades by 0.6 \% in PSNR and 0.1\% in SSIM. This evidence suggests that our SASM can reduce the computational  burden while sacrificing minimal reconstruction performance, thus increasing the model efficiency.

\begin{figure*}[t]
	\begin{center}
		\begin{tabular}[t]{c} \hspace{-2.7mm} %\vspace{-6mm}
			\includegraphics[width=\textwidth]{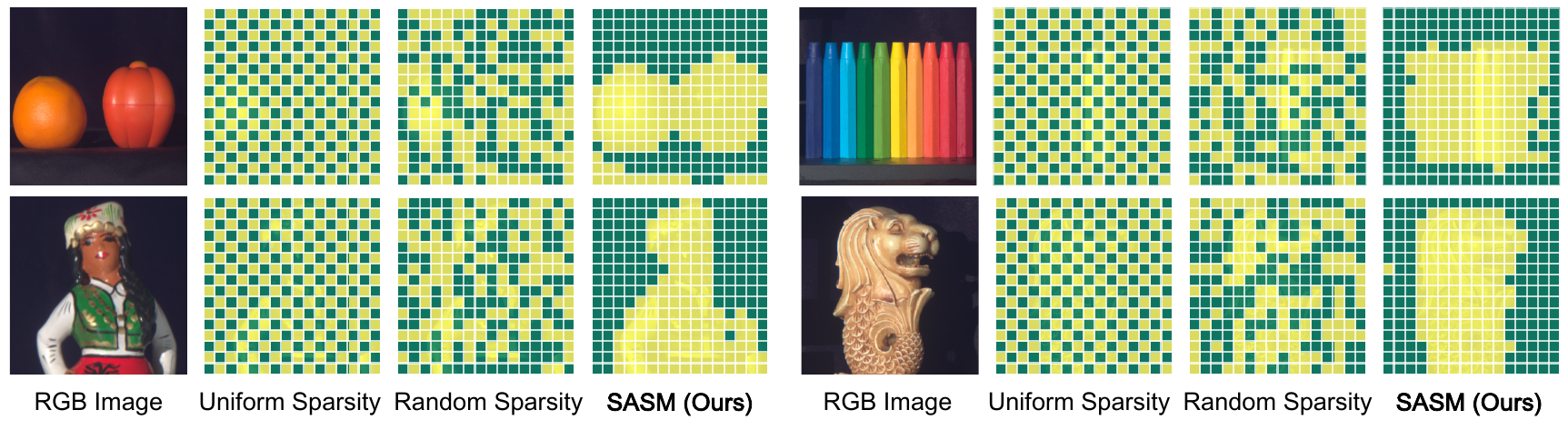}
		\end{tabular}
	\end{center}
	\vspace*{-6mm}
	\caption{\small Visual analysis of uniform sparsity scheme, random sparsity scheme, and our SASM. We visualize the sparsity masks produced by different sparsity schemes. Yellow indicates the patch is selected while green means vice versa. Only our SASM can  generate accurate response to the informative regions with HSI information.  }
	\label{fig:sparsity_mask}
	\vspace{-2mm}
\end{figure*}

\vspace{-2mm}
\subsubsection{Sparsity Scheme Comparison.} We conduct ablation to study the effects of sparsity schemes including: (i) random sparsity, \emph{i.e.}, the patches to be calculated are randomly selected, (ii)  uniform sparsity, \emph{i.e.}, the patches to be calculated are uniformly distributed, and (iii) our SASM. The results are listed in  Tab.~\ref{tab:sparse}.
Our SASM yields the best results and drastically outperforms other schemes (over 0.9 dB). Additionally, we conduct visual analysis of the sparsity mask generated by the three sparsity schemes. As depicted in Fig.~\ref{fig:sparsity_mask}, the sparsity mask produced by our SASM generates more complete and accurate responses to the informative regions with HSI information. In contrast, both random and uniform sparsity schemes are not aware of HSI signals and rigidly pick the preset positions.  These results demonstrate the superiority of our SASM in perceiving spatially sparse HSI  signals and locating regions with dense HSI representations.

\vspace{-2mm}
\subsubsection{Self-Attention Mechanism Comparison.} We compare our  SAH-MSA with other self-attention mechanisms.  The results are reported in Tab.~\ref{tab:attention}. The baseline yields 32.57 dB with 0.51 M Params and 6.40 G FLOPS. We respectively apply global MSA (G-MSA)~\cite{global_msa}, local window-based MSA (W-MSA)~\cite{liu2021swin}, Swin-MSA~\cite{liu2021swin}, spectral-wise MSA (S-MSA)~\cite{mst}, and SAH-MSA. The model gains by 2.47, 2.45, 2.55, 2.64, and 2.96 dB while adding 29.18, 18.58, 18.58, 18.34, and 18.20 G FLOPS and 1.34, 1.34, 1.34, 1.15, and 0.85 M Params. Our SAH-MSA yields the most significant improvement but requires the cheapest FLOPS and Params. Please note that we downscale the input feature of G-MSA into $\frac{1}{4}$ size to avoid memory bottlenecks. This evidence shows the cost-effectiveness advantage of SAH-MSA, which is mainly because  SAH-MSA applies MSA calculation between tokens that are closely related in content within each $bucket$ while cutting down the burden of computation between content-uncorrelated elements.

\vspace{-2mm}
\subsubsection{Clustering Scope.} We study the effect of the scope of clustering, \emph{i.e.}, local \emph{vs.} global. Local means constraining the hash clustering operation inside each $M\times M$ patch while global indicates applying the hash clustering to the whole image. In the beginning, we thought that expanding the receptive field would improve the performance. However, the experimental results in Tab.~\ref{tab:cluster} point out the opposite. The model with local clustering scope performs better. We now analyze the reason for this observation. The hash clustering is essentially a linear dimension reduction ($h:\mathbb{R}^C\to \mathbb{Z}$) suffering from limited discriminative ability. It is suitable for simple, linearly separable situations with a small number of samples. When the clustering scope is enlarged from the local patch to the global image, the number of tokens increases dramatically ($M\times M \to H\times W$). As a result, the situation becomes more complex and may be linearly inseparable. Thus, the hash clustering performance degrades. Then the elements clustered into the same $bucket$ are less content-related and the MSA calculation of each $bucket$ becomes less effective, leading to the degradation of HSI restoration.

\vspace{-2mm}
\subsubsection{Parameter Analysis.} 
\begin{figure*}[t]
	\begin{center}
		\begin{tabular}[t]{c} \hspace{-3.5mm} %\vspace{-6mm}
			\includegraphics[width=\textwidth]{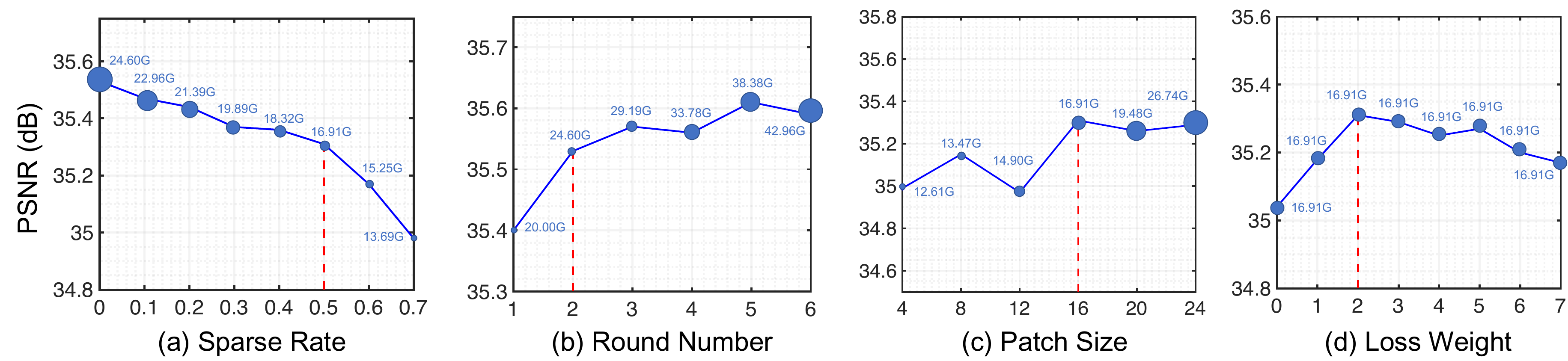}
		\end{tabular}
	\end{center}
	\vspace*{-6mm}
	\caption{\small Parameter analysis of sparsity ratio $\sigma$, round number $R$, patch size $M$, and loss weight $\lambda$. The vertical axis is PNSR (in dB performance). The circle radius is FLOPS (computational cost). To achieve a better trade-off between performance and computational complexity, we finally set $\sigma$ = 0.5, $R$ = 2, $M$ = 16, and $\lambda$ = 2.}
	\label{fig:pa}
	\vspace{-3mm}
\end{figure*}

We adopt CST-M to conduct parameter analysis of sparsity rate $\sigma$, round number $R$ in Eq.~\eqref{eq:multi_round}, patch size $M$, and loss weight $\lambda$ in Eq.~\eqref{overall_loss} as shown in Fig~\ref{fig:pa}, where the vertical axis is PSNR and the circle radius is FLOPS. As can be observed: \textbf{(i)} When increasing $\sigma$, the computational cost declines but the performance is sacrificed. When $\sigma$ is larger than $50\%$, the performance degrades dramatically. \textbf{(ii)} When  changing $R$ from 1 to 6, the reconstruction quality increases. Nonetheless, when $R \ge$ 2, further increasing $R$ does not lead to a significant improvement. \textbf{(iii)} The two maximums are achieved when $M$ = 16 and $\lambda$ = 2, respectively, without costing too much FLOPS. Since our goal is not to pursue the best results with heavy computational burden sacrificing the model efficiency but to yield a  better trade-off between performance and computational cost, we finally set $\sigma$ = 0.5, $R$ = 2, $M$ = 16, and $\lambda$ = 2. 

\vspace{-2mm}
\section{Conclusion}
\vspace{-1mm}
In this paper, we investigate a critical problem in HSI reconstruction, \emph{i.e.}, how to embed HSI sparsity into learning-based algorithms. To this end, we propose a novel Transformer-based method, named CST, for HSI restoration. CST firstly exploits SASM to detect informative regions with HSI representations. Then the detected patches are fed into our  SAH-MSA to cluster spatially scattered tokens with closely correlated contents for calculating MSA. %This coarse-to-fine learning scheme enables our CST to be more aware of the spatially sparse HSI information and reduce the inefficient computation of dark uninformative regions, thus boosting the model cost-effectiveness.  
Extensive quantitative and qualitative experiments demonstrate that our CST significantly outperforms other SOTA methods while requiring cheaper computational costs. Additionally, our CST yields more visually pleasing results with more fine-grained details and structural contents than existing algorithms in real-world HSI reconstruction.

\vspace{2mm}

\noindent \textbf{Acknowledgements:} This work is partially supported by the NSFC fund (61831 014), the Shenzhen Science and Technology Project under Grant (JSGG20210802 153150005, CJGJZD20200617102601004), and the Westlake Foundation (2021B1 501-2). Xin Yuan would like to thank the funding from Lochn Optics.

\bibliographystyle{splncs04}
\bibliography{egbib,reference}
\end{document}